\documentclass[sigconf]{acmart}

\AtBeginDocument{%
  }

\setcopyright{acmcopyright}
\copyrightyear{2018}
\acmYear{2018}
\acmDOI{XXXXXXX.XXXXXXX}

\acmConference[VRST '22]{The 28th ACM Symposium on Virtual Reality Software and Technology }{Nov. 29--Dec. 1, 2022}{Tsukuba, Japan}
\acmPrice{15.00}
\acmISBN{978-1-4503-XXXX-X/18/06}

\usepackage{bbding}
\usepackage{multirow}
\usepackage{color}

\begin{document}

\title{3D Reconstruction of Sculptures from Single Images via Unsupervised Domain Adaptation on Implicit Models}
\author{Ziyi Chang}
\email{ziyi.chang@durham.ac.uk}
\affiliation{%
  \institution{Durham University}
  \city{Durham}
  \country{UK}
}

\author{George Alex Koulieris}
\email{georgios.a.koulieris@durham.ac.uk}
\orcid{0000-0003-1610-6240}
\affiliation{%
  \institution{Durham University}
  \city{Durham}
  \country{UK}
}

\author{Hubert P. H. Shum}
\authornote{Corresponding author.}
\email{hubert.shum@durham.ac.uk}
\orcid{0000-0001-5651-6039}
\affiliation{%
  \institution{Durham University}
  \city{Durham}
  \country{UK}
}

\begin{abstract}
    Acquiring the virtual equivalent of exhibits, such as sculptures, in virtual reality (VR) museums, can be labour-intensive and sometimes infeasible. Deep learning based 3D reconstruction approaches allow us to recover 3D shapes from 2D observations, among which single-view-based approaches can reduce the need for human intervention and specialised equipment in acquiring 3D sculptures for VR museums. However, there exist two challenges when attempting to use the well-researched human reconstruction methods: limited data availability and domain shift. Considering sculptures are usually related to humans, we propose our unsupervised 3D domain adaptation method for adapting a single-view 3D implicit reconstruction model from the source (real-world humans) to the target (sculptures) domain. We have compared the generated shapes with other methods and conducted ablation studies as well as a user study to demonstrate the effectiveness of our adaptation method. We also deploy our results in a VR application.
\end{abstract}



\begin{teaserfigure}
\centering
  \includegraphics[width=0.8\textwidth]{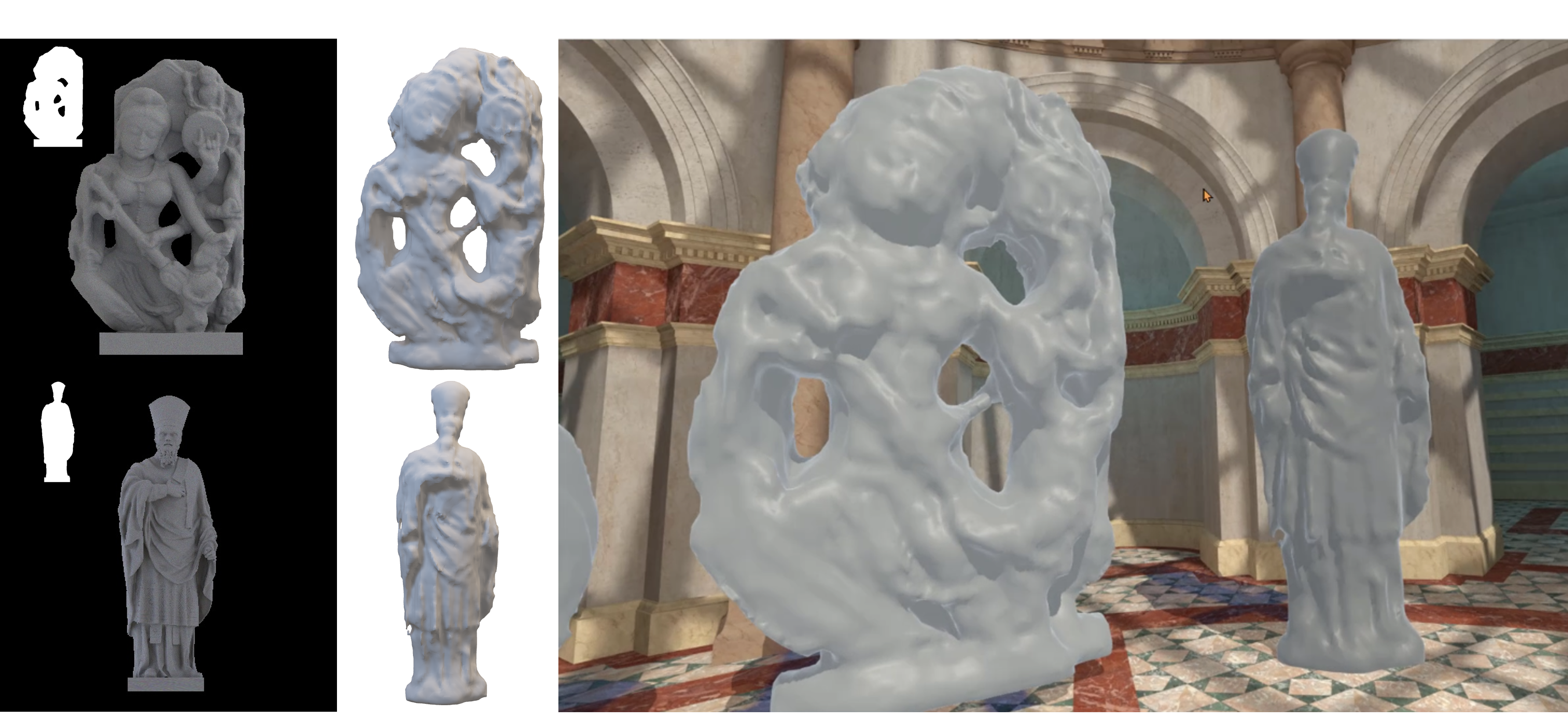}
  \caption{Two reconstruction examples (left and middle images) and display in virtual reality (right image). Our network generates 3D sculptures from single RGB images. Results can be used in virtual reality applications such as virtual museum.}
  \label{fig:teaser}
\end{teaserfigure}

\maketitle

\section{Introduction}

While virtual reality (VR) is an effective tool for implementing virtual museums, the acquisition of the virtual equivalent of exhibits, such as sculptures that we consider in this paper, can be labour-intensive. Many exhibits are 3D in nature, such as sculptures, historical objects and specimens. Typical solutions for acquiring 3D surfaces such as 3D scanning generally require specific hardware and potentially manual post-processing. A machine-learning-based, software solution that requires minimal human intervention is beneficial for VR applications. This paper focuses on sculptures as they have a wide variety of shape topology and are challenging due to the general lack of texture.


Deep learning based 3D reconstruction has the potential of effectively acquiring 3D models from 2D observations. View synthesis approaches such as NeRF \cite{mildenhall2020nerf} can render new views, but may incur inconsistency across views due to the lack of explicit 3D modelling. General object reconstructions such as \cite{onizuka2020tetratsdf} may reconstruct a wide variety of objects. However, due to the lack of prior knowledge on the type of objects, reconstruction quality may not always be consistent. 
We observe that sculptures generally resemble real-world humans while having variations on body structure. This motivates us to adapt the well-researched 3D human reconstruction methods such as \cite{saito2019pifu} to reconstruct sculptures. 

To minimise the amount of human intervention and the use of specialised equipment, implicit 3D human reconstruction from a single RGB image is preferred. Previous single-view based human reconstruction methods can be categorized into three streams: implicit reconstruction, parametric reconstruction, and their combination. The first one is to learn an implicit function such as signed distance function \cite{onizuka2020tetratsdf} and occupancy probability \cite{alldieck2022phorhum}. The second one builds upon various parametric human models such as SMPL \cite{loper2015smpl} and uses neural network to estimate a few parameters to deform the pre-defined parametric human models. The combination \cite{bhatnagar2020ipnet} blends previous two streams to add surface details to parametric human models. Compared with regular human shapes, sculptures have more irregular topology. Parametric human models cannot be deformed into another topology. To solve the irregular topology of sculptures, we consider implicit models are more suitable than parametric ones for our reconstruction problem.




There are two major challenges when adapting single-view implicit 3D reconstruction for sculptures. First, obtaining 3D data are labour-intensive and sometimes infeasible \cite{alldieck2022phorhum}, which is particularly true for sculptures as they may be of larger sizes and partially combined together with the environment. Second, sculptures generally consist of irregular topology, making it difficult to apply pre-trained models for reconstruction. This is known as domain shift \cite{zhang2021survey} in machine learning. 

Our insight is that the aforementioned challenges can be solved by unsupervised domain adaptation of 3D human reconstruction systems. To tackle the limitation of data, unsupervised learning enables our system to acquire knowledge of 3D sculptures without large well-labelled datasets \cite{musgrave2021unsupervised}. To tackle domain shift, due to the similarity of human-based shapes and sculptures, domain adaptation on an implicit model can transfer knowledge learnt from one domain into another \cite{zhang2021survey} without retraining while allowing the modelling of the different distribution of topology.


In this work, we propose a new method of unsupervised 3D domain adaptation for 3D sculpture reconstruction tasks. Our pipeline utilises single RGB images as input and transfers a pre-trained implicit model from source (real-world human) into target (sculpture) domain in an unsupervised manner. We consider pre-trained PIFu \cite{saito2019pifu} is more suitable than other implicit models due to limited data availability. 
To represent domain shift, we adapt multi-layer intermediate feature architecture and Maximum Mean Discrepancy (MMD) into our problem of 3D unsupervised domain adaptation. Multiple features in the encoder extract different patterns from low to high levels. Our MMD is based on multiple levels of features and aligns two domains. 
To refine geometry and surface details, we propose neighbour aggregation and pseudo labels for our problem of 3D unsupervised domain adaptation on 3D sculpture reconstruction. Guidance is designed to transmit from the source domain to the target domain through aggregation and pseudo labels.
During the aggregation, we re-weight features that are used to define neighbours by balancing the importance of pixel-aligned information with depth values. To the best of our knowledge, we present the first research on unsupervised 3D domain adaptation with implicit models on the 3D reconstruction task.


Experiments show our network exceeds state-of-the-art on our collected sculptures for evaluation. We have provided qualitative results generated by various methods. As for quantitative comparison, we choose Chamfer distance and point-to-surface distance as the two metrics - in both of which our methods achieve the best results. We also conduct extra experiments on other alternative domain adaptation methods to demonstrate the effectiveness of our design. Additional ablation studies on proposed modules and design choices have been implemented. In addition, we have also invited 20 volunteers to participate in our user study. Finally, we deploy the reconstructed sculptures in a VR application.

Our contributions are as follows, with source code available at {\color{blue} \url{https://github.com/mrzzy2021/SculptureRecon}}: 
\begin{itemize}
    \item We propose an unsupervised method for domain adaptation on sculpture reconstruction, solving the problem of creating 3D shapes with limited available data. 
    \item We adapt multi-level features into our problem of unsupervised 3D domain adaptation on a 3D reconstruction task.
    \item We propose to use a re-weighting and neighbourhood strategy for better structural exploration of latent feature space.
    
\end{itemize}

\section{Related Work}

\subsection{3D Reconstruction}

3D reconstruction has been applied in recovering a variety of subjects. Some approaches mainly focus on human reconstruction \cite{chen2021towards,tian2022recovering}. Others are specialized for objects \cite{han2019image,gao2019object}. For example, \cite{Corona2021SMPLicit} is designed for humans while \cite{nozawa20223d} focuses on reconstructing 3D cars. There are also methods for arbitrary subjects such as NeRF \cite{mildenhall2020nerf}. However, the reconstruction quality may not be consistent due to the lack of prior knowledge. In our paper, we observe that many sculptures are artistic interpretations of real world human beings and thus, we focus more on human reconstruction to better leverage domain knowledge and control computational cost.

Among 3D human reconstruction methods, they can be categorized into single-view and multi-view human reconstruction \cite{zheng2021pamir}. Single-view methods can recover 3D shape from a single image \cite{tian2022recovering,saito2019pifu}. Multi-view approaches make predictions based on several images or videos that carry shape information from different angles \cite{chen2021towards,hong2021stereopifu}. While multi-view human reconstruction can achieve better performance, it requires more information as input, which may not be accessible due to limited data acquisition. On the contrary, single-view reconstruction methods are more flexible and easy to operate when data acquisition is difficult. As it may not be easy to collect multi-view images for sculptures due to complex environmental restrictions, we are interested in single-view reconstruction. This also makes our methodology favorable for practical deployment of AR/VR applications among others, as it is easier to obtain a single image of a sculpture when compared with multiple images. 

There are three main streams of methods for reconstruction of real-world humans. The first stream uses implicit models, which is proposed in recent years and has become more and more popular. Implicit models rely on a learned function to describe surfaces \cite{zheng2021pamir}. There are two popular implicit representations available for implicit models. One is occupancy probability, which indicates the probability of the queried 3D coordinate inside the 3D shape and many studies have trained implicit functions to predict the probability \cite{saito2019pifu,saito2020pifuhd}. They focus on how to extract more representative image features, by aligning pixel-wise features to coordinates or iteratively extracting multi-level local features, to facilitate more accurate classification on sampled locations. The other representation is the signed distance function  \cite{onizuka2020tetratsdf,Liu2020DIST}. Instead of using probability to indicate different properties of 3D coordinates, signed distance functions indicate the positive or negative distance from sampled locations to the surface \cite{Yang2021deep}. 

The second stream of methods for the reconstruction of real-world humans uses parametric models. These models are pre-defined manually and controlled by a few parameters. A large number of parametric human models have been proposed such as SMPL \cite{loper2015smpl}, MANO \cite{romero2017embodied}, SMPL-X \cite{Pavlakos2019Expressive} and STAR \cite{Osman2020STAR}. Earlier studies usually estimate the corresponding parameters directly on the provided images. To achieve better results, recent works tend to estimate parameters in cooperation with other extra information. \cite{alldieck2019tex2shape} makes use of UV transformation to generate normals and displacements and then applies them to SMPL models to get a clothed 3D human body. \cite{weng2019photo} directly fits a SMPL model to silhouette and posture and then refines the texture of the SMPL model by normal and skinning maps. \cite{zheng2019deephuman} assigns pre-defined semantic representation to every estimated SMPL vertex and voxelises mesh to better refine shape details. \cite{Corona2021SMPLicit} proposes to use UV map to indicate occlusion and leverages convolution kernels on 2D images to encode style features and size features to model clothes which are then combined with SMPL parameters for reconstruction. 

The third stream is to combine parametric models with implicit representations for more accurate modelling \cite{zheng2021pamir}. Although in the second and third streams, parametric methods can reconstruct 3D real-world humans, they cannot handle irregular topology because of their pre-defined parametric models. Therefore, we consider implicit methods are more suitable for our reconstruction problem. 


\subsection{Domain Adaptation}

There are three types of domain adaptation methods. The first one is supervised domain adaptation \cite{liu2022end} and requires that both the source and target domains are well-labelled. The second category is semi-supervised domain adaptation \cite{chen2021semi} where partial target data are well-labelled. The third one is unsupervised domain adaptation \cite{zhang2021survey} and does not rely on target labels to transfer the learnt knowledge from the source domain to the target domain. Considering the limited data acquisition, we believe unsupervised domain adaptation is more suitable to solve our problem.

Unsupervised domain adaptation has been widely used on 2D or 2.5D tasks. Previous \textcolor{black}{researches} on 2D can be categorised into two groups. One is based on learning domain-invariant features and the other is to use a network to directly learn the mapping. The former category \cite{kang2019contrastive} minimises domain shifts by reducing the difference between feature distributions in feature space. The latter group \cite{bousmalis2017unsupervised} models the mapping by training a network such as CycleGAN \cite{zhu2017unpaired}. For 2.5D, \cite{shen2022dcl} utilises contrastive learning to maintain geometries with depth images. For these different inputs, there are a number of popular methods such as pseudo-labeling \cite{saito2017asymmetric}, and batch normalization tailored for domain adaptation \cite{maria2017autodial}.

Though lots of effort has been made on 2D images or depth maps, few works have been proposed for 3D reconstruction. Unsupervised domain adaptation for 3D reconstruction was introduced by \cite{pinheiro2019domain}. \textcolor{black}{\cite{kaya2020self} assumes their data are from the same object category and proposes a method that performs translations between 3D and 2D representations, which in other words are two domains. It focuses more on training networks by unpaired images and shapes.} \textcolor{black}{\cite{yang2021dynamic} fills the domain gap in 3D reconstruction between synthetic images and real images by extracting domain-invariant image features. This work utilises graph neural network for mesh generation and proposes DDA to extends the network to solve the synthetic-to-real difference} \cite{manamasa2020domain,qi2020unsupervised}. \textcolor{black}{Instead of synthetic-to-real differences, our research focuses on the topology shift} and we propose our domain adaptation methods to solve the sculpture reconstruction problem as a transferring problem.

\section{3D Sculpture Reconstruction}\label{sec:adaptation}
\begin{figure}[tbph]
  \centering
  \includegraphics[width=1.0\linewidth]{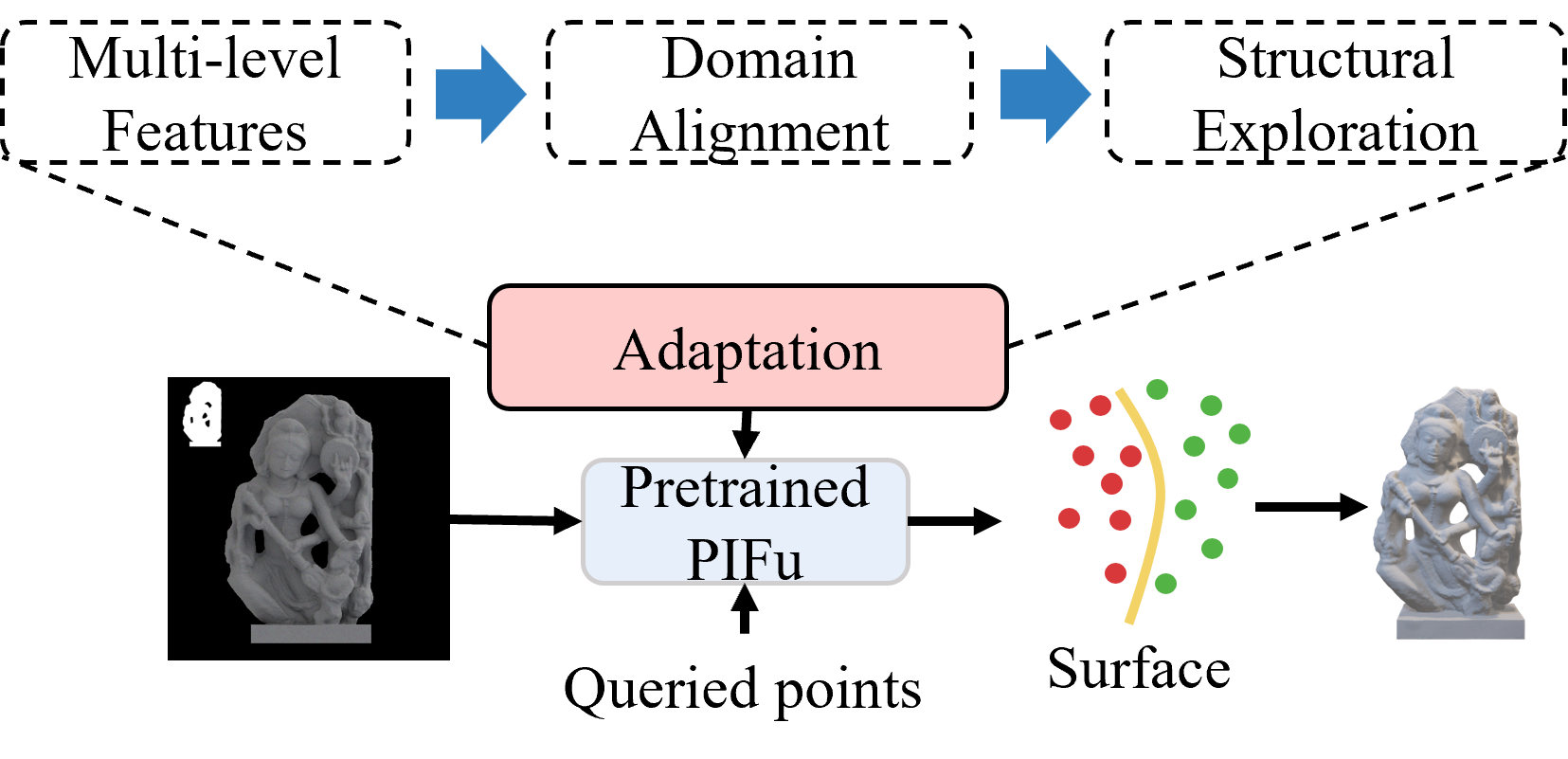}
  \caption{\label{fig:pipeline}%
           Overview of our proposed unsupervised domain adaptation on 3D sculpture reconstruction. 
           }
\end{figure}
Reconstructing 3D sculptures from single RGB images has two main challenges. First, the difficulty on data acquisition leads to limited labels and training data. It is infeasible to obtain other auxiliary information such as surface normals, prohibiting the use of some reconstruction modules \cite{saito2020pifuhd}. The other challenge is the domain shift of topology. The irregular topology of 3D sculptures makes domain transferring non-linear. The ideal 3D representation used in the reconstruction module should be flexible enough to cover various potential 3D shapes.

To solve the aforementioned challenges, we choose PIFu \cite{saito2019pifu} that is pre-trained on real-world humans and propose our unsupervised domain adaptation (UDA) pipeline. Using a human reconstruction network aids the unsupervised domain adaptation. We observe that sculptures are usually artistic interpretations of humans. To alleviate the difficulty of transferring, we leverage the already-learnt knowledge on real-world human shapes as a solid starting point. PIFu is a single-view reconstruction method and only requires a single RGB image as input, mitigating the difficulty on data acquisition. Meanwhile, PIFu is based on implicit representation that learns a function to approximate the surface. Compared with parametric representation, PIFu is not restricted by the pre-defined canonical models and is flexible enough to learn irregular topology of 3D sculptures. We also design our domain adaptation to be unsupervised to transfer knowledge of real-world humans into sculptures without requiring labels in the target domain. 



\subsection{Problem Definition}\label{sec:problemdef}

Considering the challenges, our goal is to propose an unsupervised domain adaptation (UDA) method for PIFu to solve the problem of reconstructing 3D sculptures from single RGB images. We will first briefly introduce the pre-trained PIFu used in our domain adaptation process. Then, we will discuss our unsupervised domain adaptation settings.
\begin{figure}[tbph]
  \centering
  \includegraphics[width=0.85\linewidth]{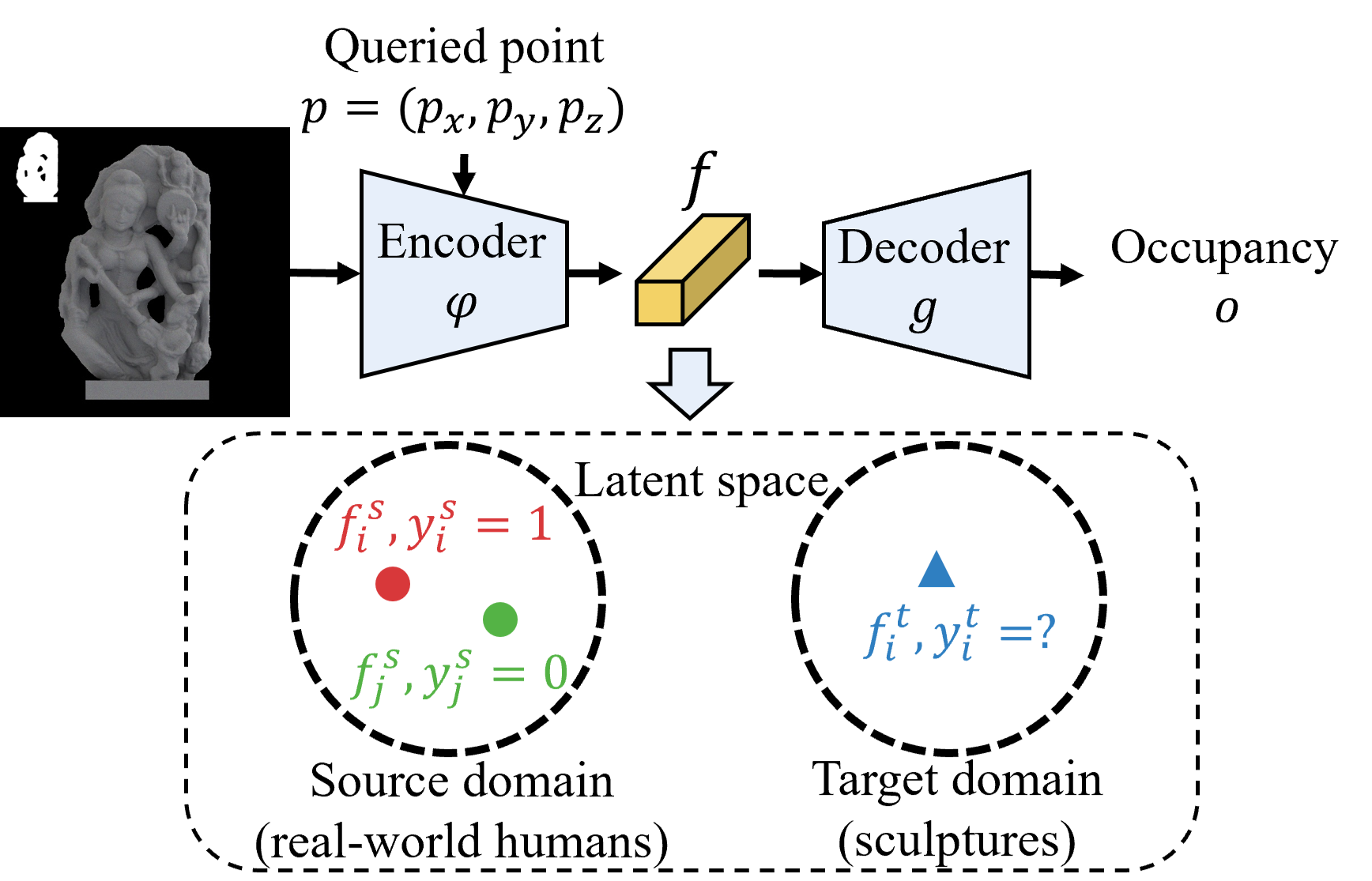}
  \caption{\label{fig:problemdef}%
           Domain definition. Our source and target domains are composed of pixel-aligned features for each queried point. The target domain only has features $f^t$ from sculptures and does not have ground truth $y^t$. 
           }
\end{figure}
The pipeline of PIFu is shown in Figure \ref{fig:problemdef}. PIFu predicts the occupancy probability for any queried point in space. It focuses on learning an implicit function $func(p,I,M)$ for a query point $p=(p_x,p_y,p_z)\in \mathbb{R}^3$, which is formulated as:
\begin{equation}\label{eq:pifu}
func(p,I,M) = g(\varphi(p_x,p_y,I,M),p_z) = o\in [0,1],
\end{equation}
where $I$ and $M$ are an image and its mask, $\varphi$ and $g$ are encoder and decoder, respectively, $o$ is the predicted occupancy probability for the queried point $p$. 
The pixel-aligned feature $f$ in PIFu for the queried point $\textcolor{black}{p=(p_x,p_y,p_z)}$ is formed by two concatenated parts: a feature based on bilinear interpolation with $(p_x,p_y)$ and the corresponding depth value $p_z$. 
The learned shape $\mathcal{S}$ is extracted using marching cubes: 
\begin{equation}
    \mathcal{S}=\{p\in \mathbb{R}^3|o=0.5\}.
\end{equation}

Our 3D unsupervised domain adaptation (UDA) settings are based on the pixel-aligned features extracted by the PIFu encoder. 
For our UDA with implicit reconstruction models, we have access to the labeled source domain that consists of $n_s$ queried points for each real-world human mesh. The PIFu encoder then converts these queried points into their corresponding pixel-aligned features. As we use multi-level architecture (to be explained in Section \ref{sec:multi-level}), we define our source domain as $\mathcal{D}^{s}=\{f_i^{s,l},y_i^s\}$ for $i=1,...,n_s$ and $l=1,...,L$ where the source labels $y_i^s$ are defined as:
\begin{equation}\label{eq:label}
    y_i^s=  \begin{cases}
                1 & p_i^s\ is\ inside\ mesh\ surface,\\
                0 & otherwise,\\
            \end{cases}
\end{equation}
and $n_s$ is the number of pixel-aligned features. Similarly, the target domain for the $l$-th layer is defined as $\mathcal{D}^{t,l}=\{f_i^{t,l}\}$ for $i=1,...,n_t$, which consist of $n_t$ features and the target labels are unknown. Our aim is to estimate their corresponding occupancy probability $o^{t,l}\in \mathcal{Y}^{t,l}$ through our UDA method. \textcolor{black}{Note that $y$ in Eq.\ref{eq:label} represents the ground truth label, which is only possibly 1 or 0. On the contrary, $o$ in Eq.\ref{eq:pifu} means the predictions of our neural network, which is within the range from 0 to 1.}

\subsection{Multi-level Features}\label{sec:multi-level}
\begin{figure}[tbph]
  \centering
  \includegraphics[width=0.95\linewidth]{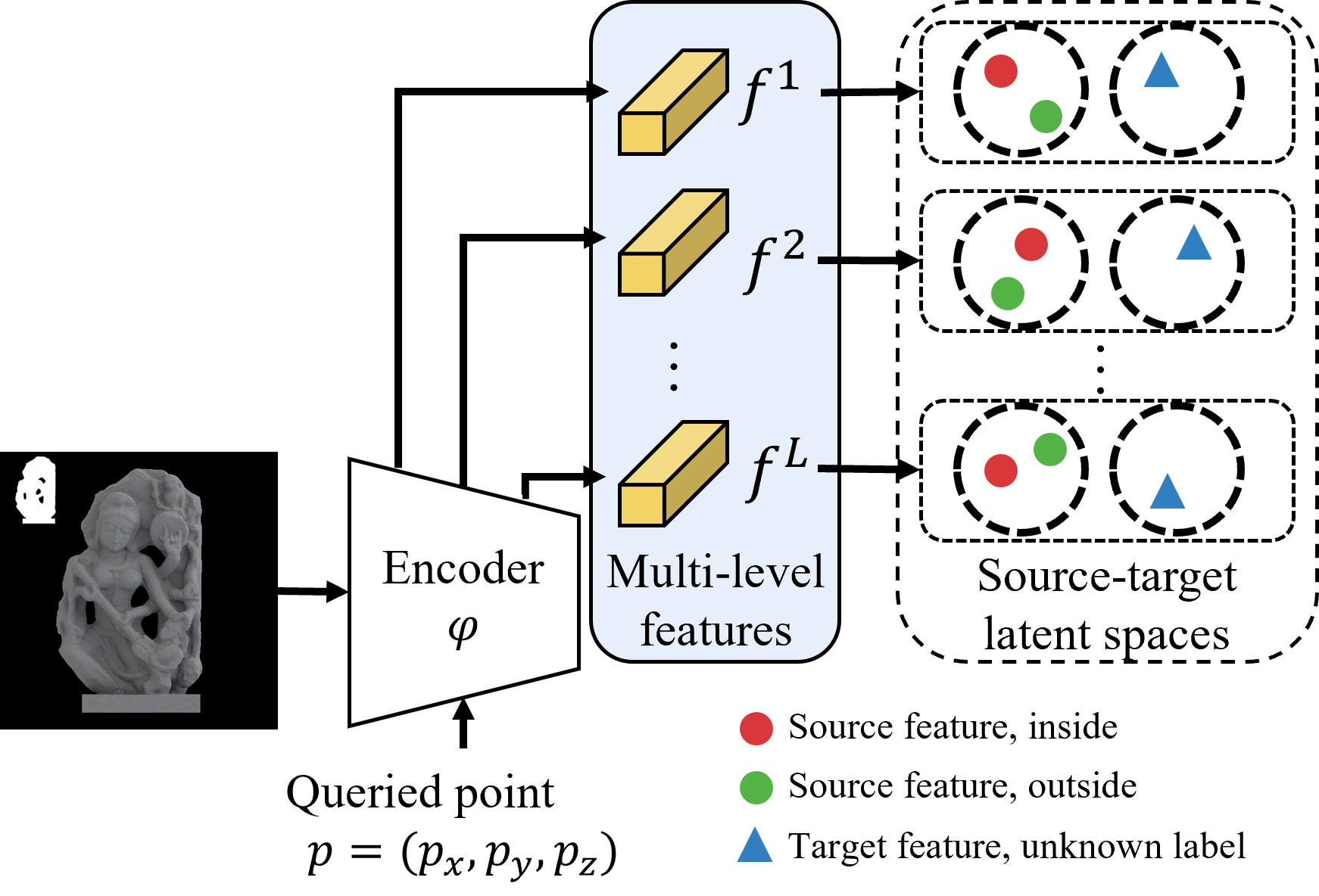}
  \caption{\label{fig:multi-level}%
           We propose to adapt multi-level pixel-aligned features to solve our unsupervised domain adaptation problem.
           }
\end{figure}

We propose to adapt the multi-level pixel-aligned features into our UDA method for the sculpture reconstruction problem, as shown in Figure \ref{fig:multi-level}. The latent features from different encoder layers are usually sensitive to a subset of input patterns. This facilitates our UDA to better measure domain shifts using comprehensive information from low to high network levels.

We also propose to process the multi-level features layer by layer dynamically. Instead of mixing them up, our design processes these features layer by layer. In this way, the relative information of occupancy within each layer can be maintained and perceived by our adaptation method. 

The multi-level pixel-aligned features come from image features extracted by the PIFu encoder. We choose four output layers of the stacked hourglass network in the PIFu encoder to obtain the pixel-aligned features. As shown in Figure \ref{fig:multi-level}, we first extract the multiple image features by different layers of PIFu encoder and then we assign the pixel-aligned features to each queried point according to their pixel-aligned 2D locations. After obtaining features, we form a number of source-target latent spaces with features from the same layers. Multi-layer features $\{f^l\}_{l=1}^L$ are extracted when given a point $p$ and layer $l$.


\subsection{Domain Alignment}\label{sec:align}
We propose our domain alignment method to align features from the source (real-world humans) and the target (sculpture) domains for our reconstruction task. Our domain alignment has two components, minimising discrepancy and maintaining accuracy.

\begin{figure}[tbph]
  \centering
  \includegraphics[width=1.0\linewidth]{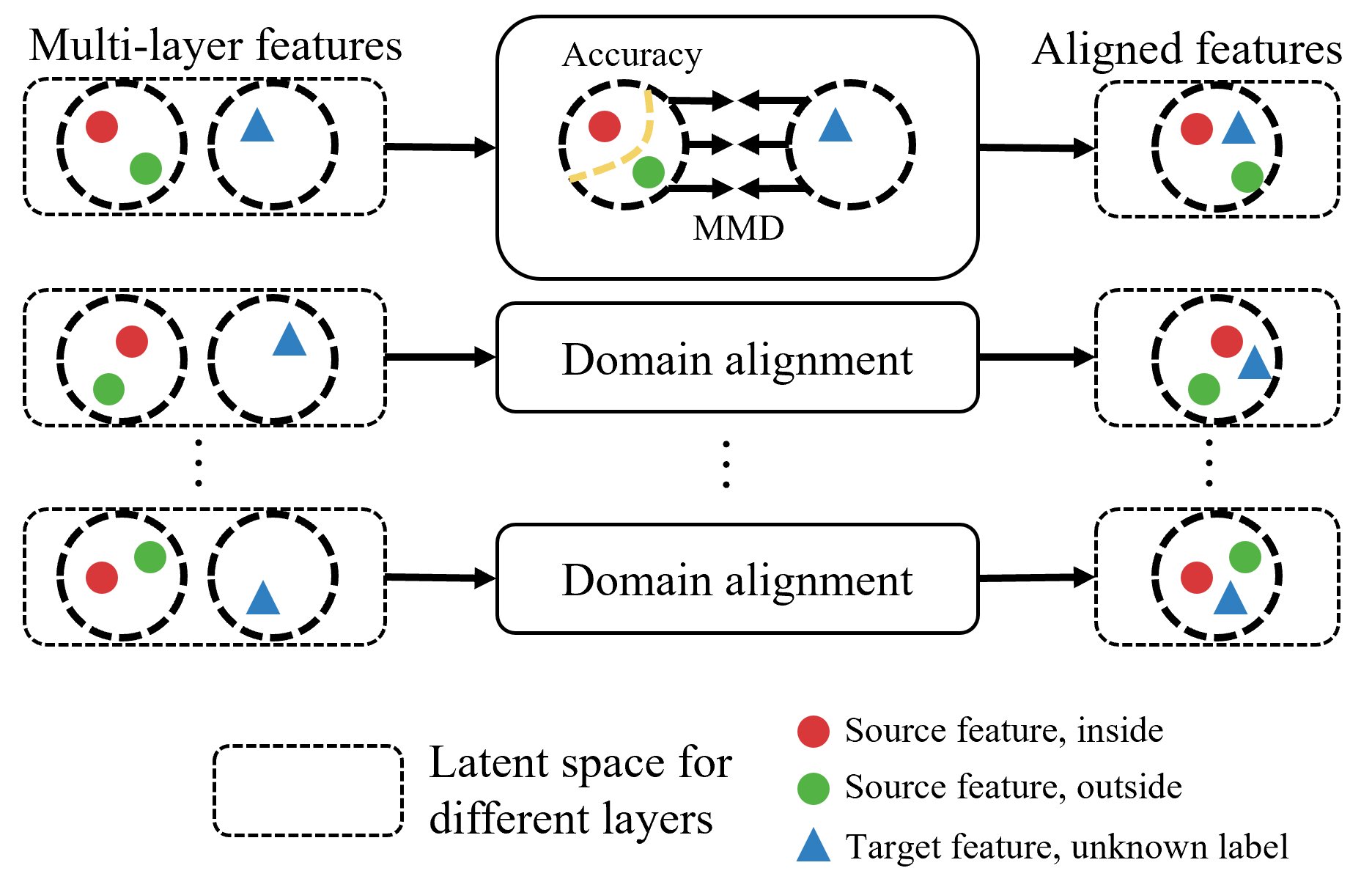}
  \caption{\label{fig:align}%
           Our domain alignment utilises Maximum Mean Discrepancy and maintains the accuracy in the source domain. 
           }
\end{figure}

For the first component, we propose to adapt Maximum Mean Discrepancy (MMD) \cite{granger2020joint} on pixel-aligned features from each layer to mix up the source and target domains for 3D reconstruction. MMD is a predefined kernel metric to quantify the similarity between two distributions. It leverages kernel functions to map pixel-aligned features from two domains into a latent space where the similarity can be calculated.

Compared to other learning-based methods, MMD is more suitable due to the limited data availability. As discussed in Section \ref{sec:problemdef}, data acquisition is a challenge that leads to limited training data without labels. Under this problem scenario, we choose to use this predefined measurement instead of learnt-based methods such as discriminators in adversarial training. Although learnt-based methods are more flexible, they require abundant training data to extract knowledge for obtaining appropriate representations. For example, discriminators trained on our small dataset cannot acquire enough knowledge to provide guidance for adversarial learning, which we will show results for in Section \ref{sec:experiments}. By decreasing the distribution difference, the encoder is forced to pay more attention to occupancy-related features.

Following our multi-layer architecture, our MMD is imposed to pixel-aligned features for each layer and we use the average value as our similarity loss for the adaptation process. Each pixel-aligned feature is projected into the Reproducing Kernel Hilbert Space (RKHS). The empirical expectation is computed for two domains and the difference can also be computed. As there are several MMD values from multiple layers, we use the average value of the computed MMD values across layers as the similarity loss. It can be formulated as:
\begin{equation}
    \begin{split}
        \mathcal{L}_{sim}&=\frac{1}{L}\sum_{l=1}^L{\mathcal{L}_{mmd}^l},\\
        &=\frac{1}{L}\sum_{l=1}^L{||\frac{1}{n_s}\sum_{f^{s,l}\in \mathcal{D}^{s,l}}{f^{s,l}} - \frac{1}{n_t}\sum_{f^{t,l}\in \mathcal{D}^{t,l}}{f^{t,l}}||_\mathcal{H}^2},
    \end{split}
\end{equation}
where $\mathcal{H}$ corresponds to the Reproducing Kernel Hilbert Space (RKHS) with Gaussian kernel. 

The second component is designed to keep accuracy on predictions in the source domain. The domain aligning operation changes pixel-aligned features to achieve domain invariance. As only the similarity is considered in MMD values, we propose to keep the accuracy of occupancy prediction in source domain where points are well-labelled. We regulate the aligned features to be meaningful by leveraging the mean square error (MSE) to maintain classification accuracy with labels $y^s\in \{0,1\}$:
\begin{equation}
    \mathcal{L}_{source}=MSE(g(f^{s,l}),y^s).
\end{equation}


\subsection{Latent Structural Exploration}

Due to the lack of target labels, we propose our method to explore the spatial structure of latent feature space with guidance from the source domain labels. Our exploration consists of two components, neighbour aggregation and diversity maintenance.

\begin{figure}[tbph]
  \centering
  \includegraphics[width=0.75\linewidth]{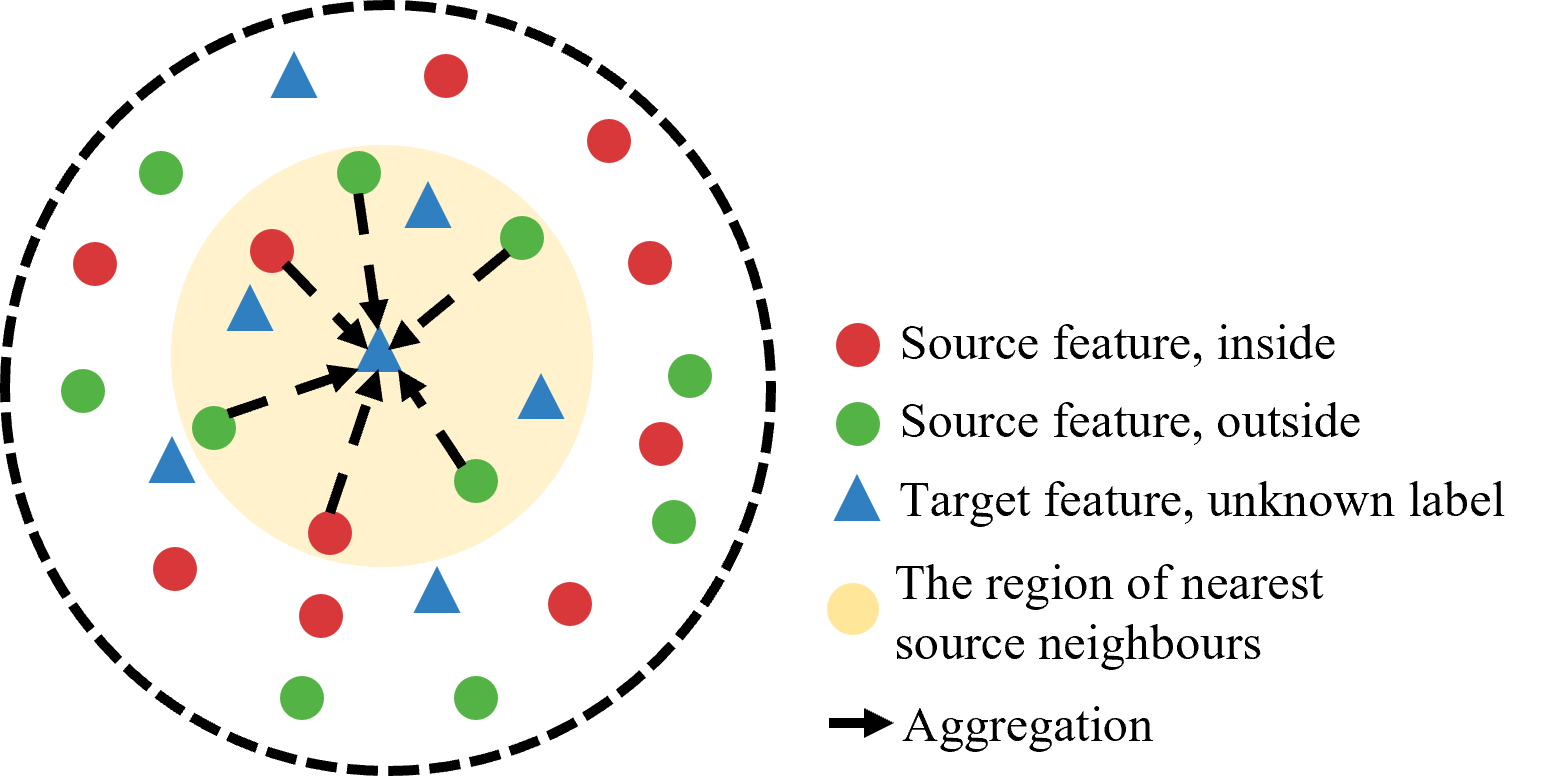}
  \caption{\label{fig:aggregation}%
           We propose to use the averaged labels from source neighbours as the indication of regional density.
           }
\end{figure}

As shown in Figure \ref{fig:aggregation}, the first component, neighbour aggregation, is designed to describe region density by averaging source neighbour labels and updating the target domain with pseudo labels. Generally, the higher region density a particular label has, the higher possibility that a point in this region should have the same label. Considering the noise of target predictions, we treat the top-K nearest features from only the source domain as neighbours. The averaged source neighbour label has been used to combine with the target predictions as the pseudo labels. This combination transmits the density clue provided by source labels to the target domain.

We empirically propose to increase the importance of depth value in features within neighbour aggregation. As single-view 3D reconstruction is an ill-posed problem and usually suffers from depth ambiguity, we hereby increase the importance of depth value in queried point coordinates when searching neighbours. Specifically, we re-weight pixel-aligned features by multiplying a scaling factor to the depth value which we empirically set as 256. The re-weighted features can better reveal structure and relationship when measured by Euclidean distance.

As we are using a multi-layer architecture, we calculate the neighbourhood information for each layer. Concretely, we use the averaged labels over source domain neighbours to represent the region density for target domain points. The aggregated information for point-feature $f^{t,l}$ from the $l$-th layer is defined as:
\begin{equation}
    \hat{y}^{t,l}=\frac{1}{|\mathcal{N}_k^{s,l}|}\sum_{f^{s,l}\in\mathcal{N}_k^{s,l}}{y^{s,l}},
\end{equation}
where $\hat{y}^{t,l}$ is the aggregated information for the target intermediate feature $f^{t,l}$ in $l$-th layer and $\mathcal{N}_k^{s,l}$ is the set of top-K nearest source features to $f^{t,l}$ in the latent space.



Our pseudo labels are designed to combine source regional density and target domain predictions. Instead of directly assigning the computed neighbourhood information as the pseudo labels for target domain points, we consider the characteristics of target structure. A momentum is used to balance the combination of target and source structural information. The aggregated result $\hat{y}^{t,l}$ is used for assigning pseudo-labels to target feature points. These labels convey information regarding the surroundings of a feature point and thus can provide guidance on target domain learning. 
The assigned pseudo-label can be calculated by balancing the self-momentum and its neighbourhood information:
\begin{equation}
    \hat{y}^{t,l} \leftarrow m\times o^{t,l} + (1-m)\times\hat{y}^{t,l},
\end{equation}
where $o^{t,l}$ is original prediction for target domain, $m$ is the self-momentum and increases as $m=\frac{epoch-start\_epoch}{epoch\_total}$
where we set $start\_epoch=30$ and $epoch\_total=60$ due to larger noise for self-predictions at the beginning of the process. 
The assigned pseudo-label is then used in MSE loss on the target domain:
\begin{equation}
    \mathcal{L}_{target}=MSE(g(f^{t,l})),\hat{y}^{t,l}).
\end{equation}

The second component is concerned with prediction diversity of the target domain by maximizing the mutual information loss. As discussed in Section \ref{sec:multi-level}, implicit-based reconstruction requires to sample points from the entire space to represent a 3D shape. \textcolor{black}{As we are moving point-features in high-dimensional latent space, the diversity is maintained to avoid the potential overly-concentration in a small area.} Therefore, we compute the entropy-based loss as our diversity measurement:
\begin{equation}
	\begin{split}
	\mathcal{L}_{mi} &= H(\mathcal{Y}^{t,l})-H(\mathcal{Y}^{t,l}|\mathcal{D}^{t,l}),\\
	                 &= h(\mathbb{E}_{f^{t,l}\in \mathcal{D}^t}(g(f^{t,l})))-\mathbb{E}_{f^{t,l}\in \mathcal{D}^t}(h(g(f^{t,l}))),
	\end{split}
\end{equation}
where \textcolor{black}{$\mathbb{E}$ is the expectation over point-features and} $h(x)=-\sum_i{xlogx}$ is the conditional entropy.

Combining the aforementioned losses, we design the overall loss as a weighted sum:
\begin{equation}
    \mathcal{L}=w_1\mathcal{L}_{sim}+w_2\mathcal{L}_{source}+ w_3\mathcal{L}_{target} + w_4\mathcal{L}_{mi},
\end{equation}
where $\{w_i\}_{i=1}^4$ are weights for different loss elements and are empirically set to $w_1=5$, $w_2=2$. We use dynamical values for $w_3$ and $w_4$ as $\frac{epoch - start\_epoch}{epoch\_total}$ where we define $start\_epoch=30$ and $epoch\_total=60$. This is because $\mathcal{L}_{target}$ will be inaccurate and diversity will be less important at the beginning of training.

\section{Experiments}\label{sec:experiments}

\subsection{Implementation Details}

\textcolor{black}{As previously mentioned, obtaining 3D data of sculptures is non-trivial and we want to figure out whether our method can work with limited data size. We collected 28 sculpture meshes from the website \textit{ScanTheWorld} and split them into a training set (20 meshes) and a testing set (8 meshes). The testing set has been hidden from all methods, including supervised and unsupervised ones, in their training process. We report quantitative performance on the testing set. As for real-world humans, we keep the same setting of what PIFu uses. We found that using only 3 real-world human meshes works for our method as we want to reduce the dependence on large amounts of labelled data required in our domain adaptation.} As the collected sculptures do not have colour, we assign random colours to each vertex. Both sculptures and real-world humans are processed to obtain RGB images and corresponding masks under the same rendering settings as used in PIFu \cite{saito2019pifu}.


\begin{figure*}[tbph]
  \centering
  \includegraphics[width=0.975\linewidth]{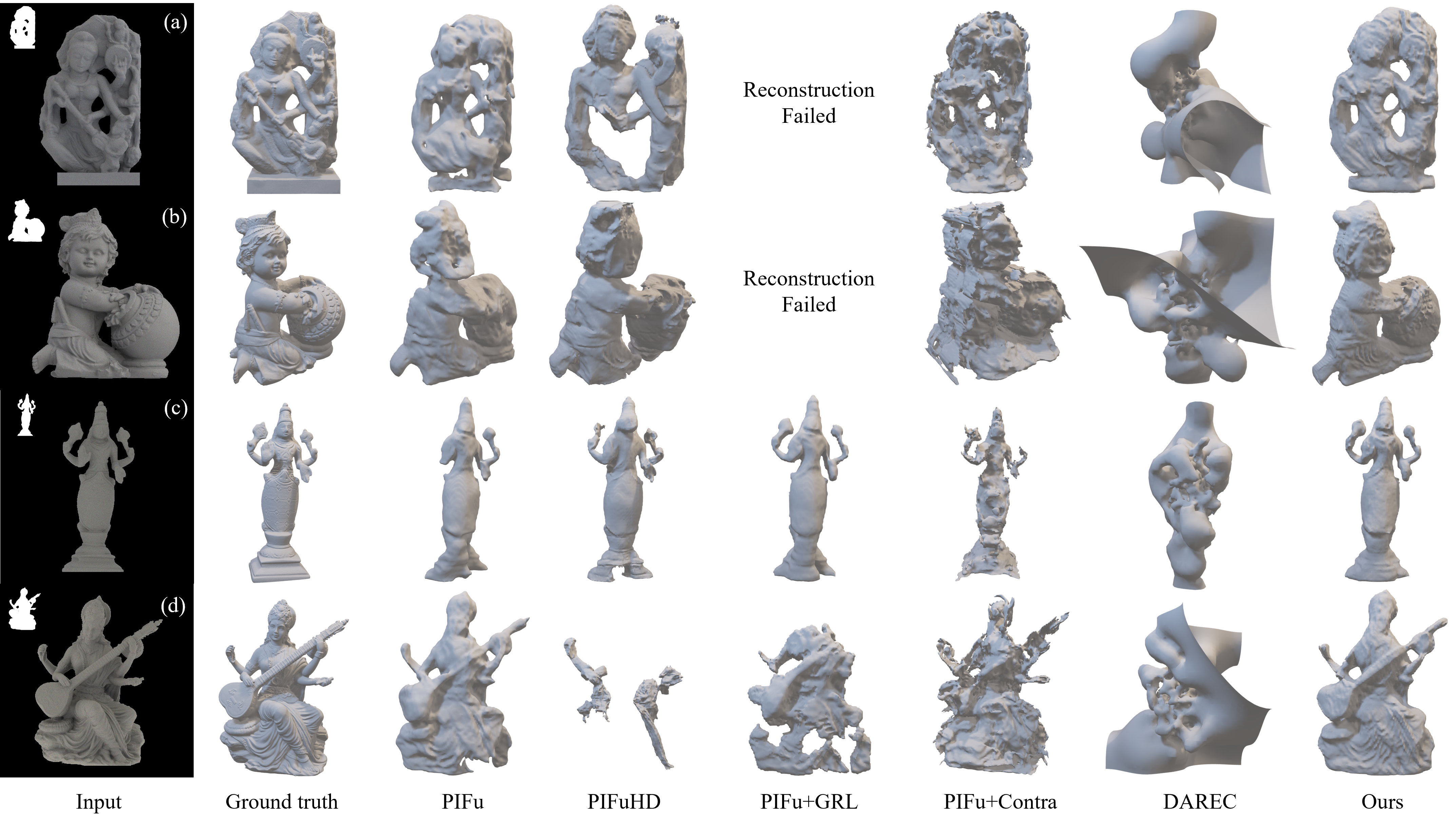}
  \caption{\label{fig:qualitative}%
          Qualitative comparison. DAREC produces point clouds and we transfer them into meshes.
          }
\end{figure*}

We use the point-to-surface distance (P2S) and chamfer distance (CD) for the quantitative performance evaluation. The former captures the surface details by computing the distance between points in the reconstructed surface and their nearest surface of ground truth. The latter metric captures the overall similarity of global shapes by computing the distance between points from the reconstructed shape and ground truth. The smaller both metrics are, the higher the quality is.

We run our code on one TITAN XP GPU. The training can be completed within approximately 2 days. We use the Marching Cubes algorithm \cite{lorensen1987marching} to extract the 0.5-level surface as the mesh shape. We filter isolated parts, rescale meshes into a unit bounding box, and place the mesh in the origin of the coordinate system.

\subsection{Quantitative and Qualitative Comparisons}

\begin{figure}[tbph]
  \centering
  \includegraphics[width=\linewidth]{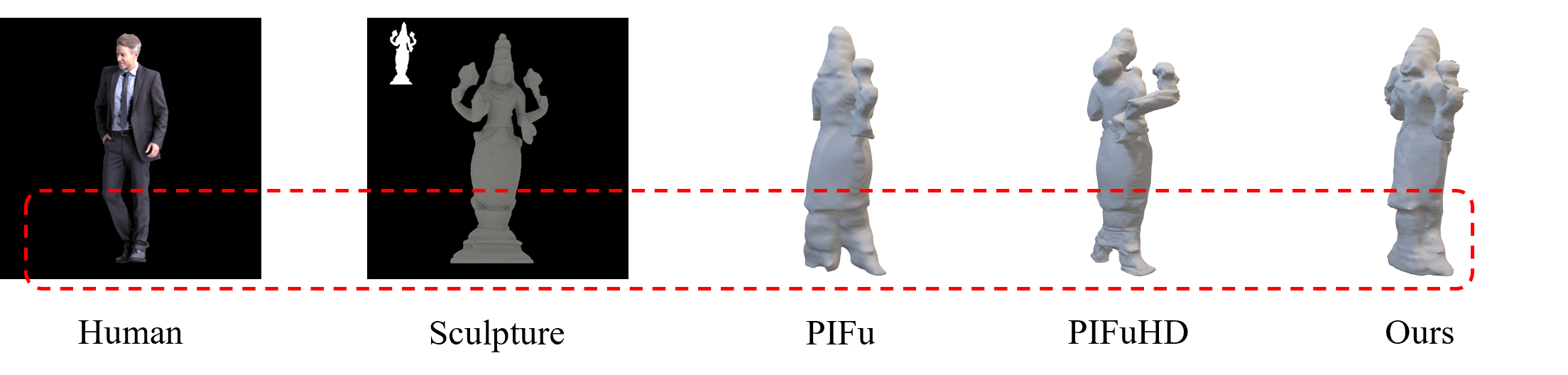}
  \caption{\label{fig:topology}%
          \textcolor{black}{An example of topology change. Both PIFu and PIFuHD tend to mimic the real-world human shape distribution, such as trying to reconstruct two legs and feet.}
          }
\end{figure}

We firstly compare our method with untrained PIFu \cite{saito2019pifu} and untrained PIFuHD \cite{saito2020pifuhd} as baseline. Both PIFu and PIFuHD are pre-trained on real-world humans and our method is learnt from our training set in an unsupervised manner. All the three compared methods are not exposed to 3D ground truth. The quantitative performance is reported on our testing set as shown in Table \ref{table:quantitative}. Our method surpasses the other two methods in both metrics. The qualitative comparison is shown in Figure \ref{fig:qualitative}. Our method reconstructs complete shapes and \textcolor{black}{corrects} surface details. \textcolor{black}{When trained with the shape of real-world humans, the model only learns a canonical human shape. Therefore, the model attempts to imitate the shape of a real-world human body for 3D reconstruction. As shown in Figure \ref{fig:topology}, PIFu and PIFuHD tend to recover two legs and feet that are usually observed in the lower part of the real-world human shapes while the sculpture has a base instead of human legs and feet. Such a topology change can be handled by our method.}

\begin{table}[tbph]
\centering
\caption{Quantitative comparison between different methods. All methods compared here do not have access to 3D ground truth of sculptures. We use $\dagger$ to denote a method fails on some results after filtering.
}
\label{table:quantitative}
\begin{tabular}{lll}
\hline
Methods& P2S& CD\\
\hline
PIFuHD \cite{saito2020pifuhd}&0.084$^\dagger$&0.099$^\dagger$\\
PIFu \cite{saito2019pifu}    &0.044&0.048\\
\hline
PIFu \cite{saito2019pifu} + Contra \cite{chen2020simple}&0.080&0.057\\
PIFu \cite{saito2019pifu} + GRL \cite{ganin2016domain}&0.06$^\dagger$&0.069$^\dagger$\\
\textcolor{black}{DAREC \cite{pinheiro2019domain}}&\textcolor{black}{0.106}&\textcolor{black}{0.087}\\
\hline
Ours                        &\textbf{0.038}&\textbf{0.047}\\
\end{tabular}
\end{table}

\begin{figure}[tbph]
  \centering
  \includegraphics[width=0.8\linewidth]{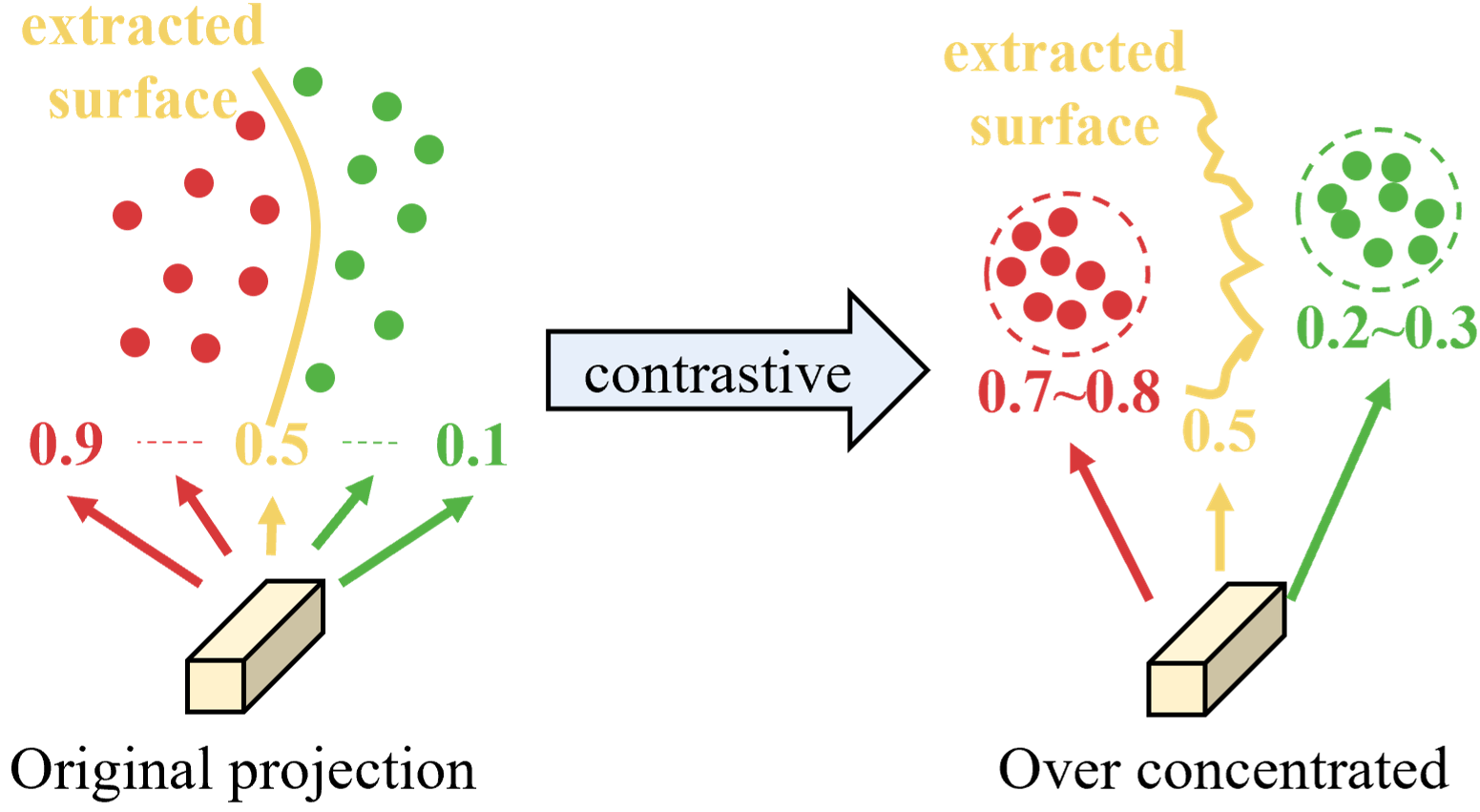}
  \caption{\label{fig:contra}%
           Explanation for `PIFu + Contra' on its over-concentrated feature space. 
           }
\end{figure}

We then compare our method with alternative unsupervised adaptation methods to validate our unsupervised design. As no previous study is conducted on unsupervised domain adaptation for 3D reconstruction with implicit models, we reimplement two adaptation methods: contrastive learning (Contra) based on \cite{chen2020simple} and adversarial learning with gradient reversal layer (GRL) based on \cite{ganin2016domain}. We train them with PIFu (denoted as PIFu + Contra and PIFu + GRL) on our training set in an unsupervised way. Table \ref{table:quantitative} and Figure \ref{fig:qualitative} show the quantitative and qualitative results, respectively. We also implement DAREC \cite{pinheiro2019domain}.
PIFu + GRL does not perform well because the discriminator cannot provide correct guidance due to the limited dataset size. \textcolor{black}{\cite{pinheiro2019domain} proposes a domain adaptation method, DAREC, to solve synthetic-to-real 3D object reconstruction. Their method leverages a 3D shape autoencoder that is pretrained on a large amount of well-labelled 3D shapes. Then an image encoder is trained with two discriminators to achieve domain confusion between extracted features from images and keep extracted features lying on the shape manifold. There are two main differences between \cite{pinheiro2019domain} and ours. The first one is that \cite{pinheiro2019domain} manipulates the single global shape-features to learn canonical human shapes for domain confusion while ours uses a set of point-features to learn occupancy probability for domain confusion. The second one is that \cite{pinheiro2019domain} relies on discriminators for domain confusion while ours is based on predefined MMD. The two differences enable our method to complete the domain adaptation in our research while \cite{pinheiro2019domain} does not perform well. The manifold formed by shape-features is based on whether discriminators consider reasonable and thus, cannot handle topology changes. Our point-features form the manifold determined by whether the point is inside or not. This provides flexibility on topology changes. Similar with PIFu + GRL, the discriminators in DAREC for domain confusion require a large amount of training data and cannot provide correct guidance due to the limited dataset size. On the contrary, our predefined MMD-based method can achieve domain confusion.} PIFu + Contra does not perform well because implicit models rely on samples around 0.5 to approximate where the 0.5-level surface is. Contrastive learning reduces the distance of the similar samples in the latent space. The concentration to some particular values leads to large uncertainty around 0.5-level surface as shown in Figure \ref{fig:contra}.

\begin{figure}[tbph]
  \centering
  \includegraphics[width=\linewidth]{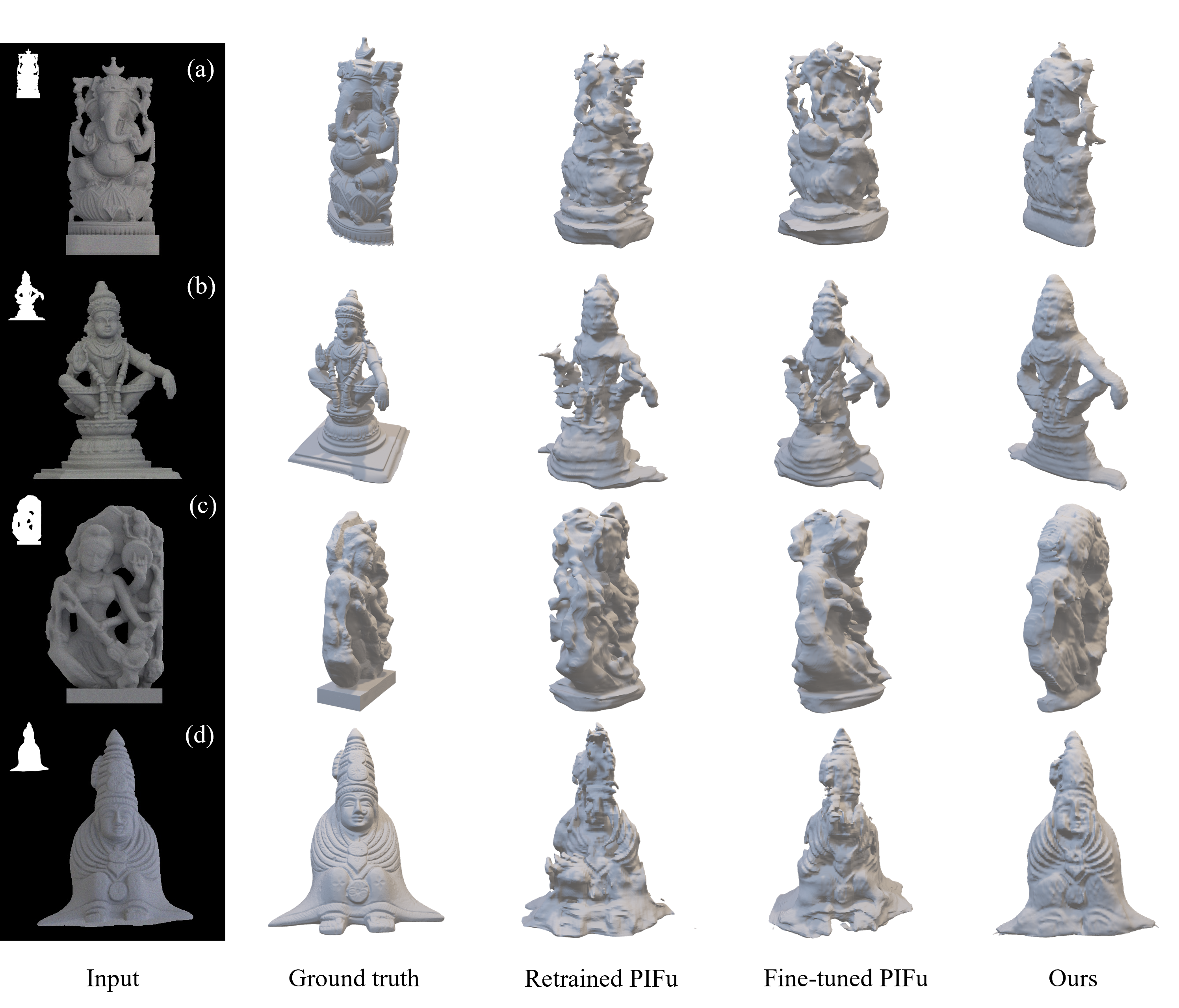}
  \caption{\label{fig:supervised}%
          Qualitative comparison with supervised methods, retrained PIFu and fine-tuned PIFu. The two methods they severely struggle with surfaces.}
\end{figure}

We compare our unsupervised method with two supervised PIFu methods. We train the two supervised methods, denoted as retrained PIFu and fine-tuned PIFu, with 3D labels in our training set. The two supervised methods directly learn with 3D ground truth labels in our training set. The testing set is hidden during training. These unseen data are used to report the quantitative performance in this experiment. Table \ref{table:quantitative2} shows quantitative results. Our method achieves the best P2S but does not in CD. However, CD is not completely consistent with visual quality \cite{jin2020dr} and thus, we provide a qualitative comparison in Figure \ref{fig:supervised}. We find out that both retrained PIFu and fine-tuned PIFu recover noisy surface.

\begin{table}[tbph]
\centering
\caption{Quantitative comparison between our unsupervised method with two supervised methods.
}
\label{table:quantitative2}
\begin{tabular}{ccc}
\hline
Methods& P2S& CD\\
\hline
Retrained PIFu&0.048&\textbf{0.039}\\
Fine-tuned PIFu&0.046&\textbf{0.039}\\
\hline
Ours        &\textbf{0.038}&0.047\\
\end{tabular}
\end{table}

The two supervised methods have a severe generalization issue and, as a result, they cannot be used in real-world applications. Figure \ref{fig:supervised} shows they do not perform well on testing data although their CD is better than ours. As pointed out by \cite{jin2020dr}, this is because CD is relatively insensitive to visual artifacts such as poor surface quality. This is also the reason that we use P2S to capture surface quality and visualize the reconstructed results. This generalization problem is caused by insufficiency of the training data and thus, retraining or fine-tuning is incapable of learning surface.


\subsection{Ablation study}

Our ablation study validates \textcolor{black}{four} modules (indicated by loss term) used in domain adaptation. We provide results under different modules used in domain adaptation in Table \ref{table:analyseloss}. 

\begin{table}[tbph]
\centering
\caption{Ablation study. $\ddag$ means a method fails to reconstruct meshes after filtering.}
\label{table:analyseloss}
\begin{tabular}{lll}
\hline
Methods&P2S&CD\\ \hline
\textcolor{black}{Ours w/o $\mathcal{L}_{mmd}$}&\textcolor{black}{0.053}&\textcolor{black}{0.052}\\
Ours w/o $\mathcal{L}_{source}$&$\ddag$&$\ddag$\\
Ours w/o $\mathcal{L}_{target}$&0.041&0.049\\
\textcolor{black}{Ours w/o $\mathcal{L}_{mi}$}&\textcolor{black}{0.041}&\textcolor{black}{0.047}\\
\hline
Ours w/o multi-level features&0.056&0.052\\
Ours w/o rescaling&0.043&0.05\\
\hline
Ours&\textbf{0.038}&\textbf{0.047}\\
\end{tabular}
\end{table}

We also provide a quantitative comparison by disabling two design \textcolor{black}{choices:} the multi-level features and the rescaling factor. Without a multi-level architecture for unsupervised adaptation, our performance drops a lot on both metrics. This indicates that multi-level features provide more comprehensive clues for our unsupervised domain adaptation. We also find that performance improves when we use a rescaling factor to balance the importance of depth value and pixel-aligned features. With this rescaling factor, our adaptation leverages the latent structure in feature space better. As shown in Table \ref{table:analyseloss}, by applying two choices, our method can achieve the best performance.

\subsection{User study}

\begin{figure}[tbph]
  \centering
  \includegraphics[width=0.7\linewidth]{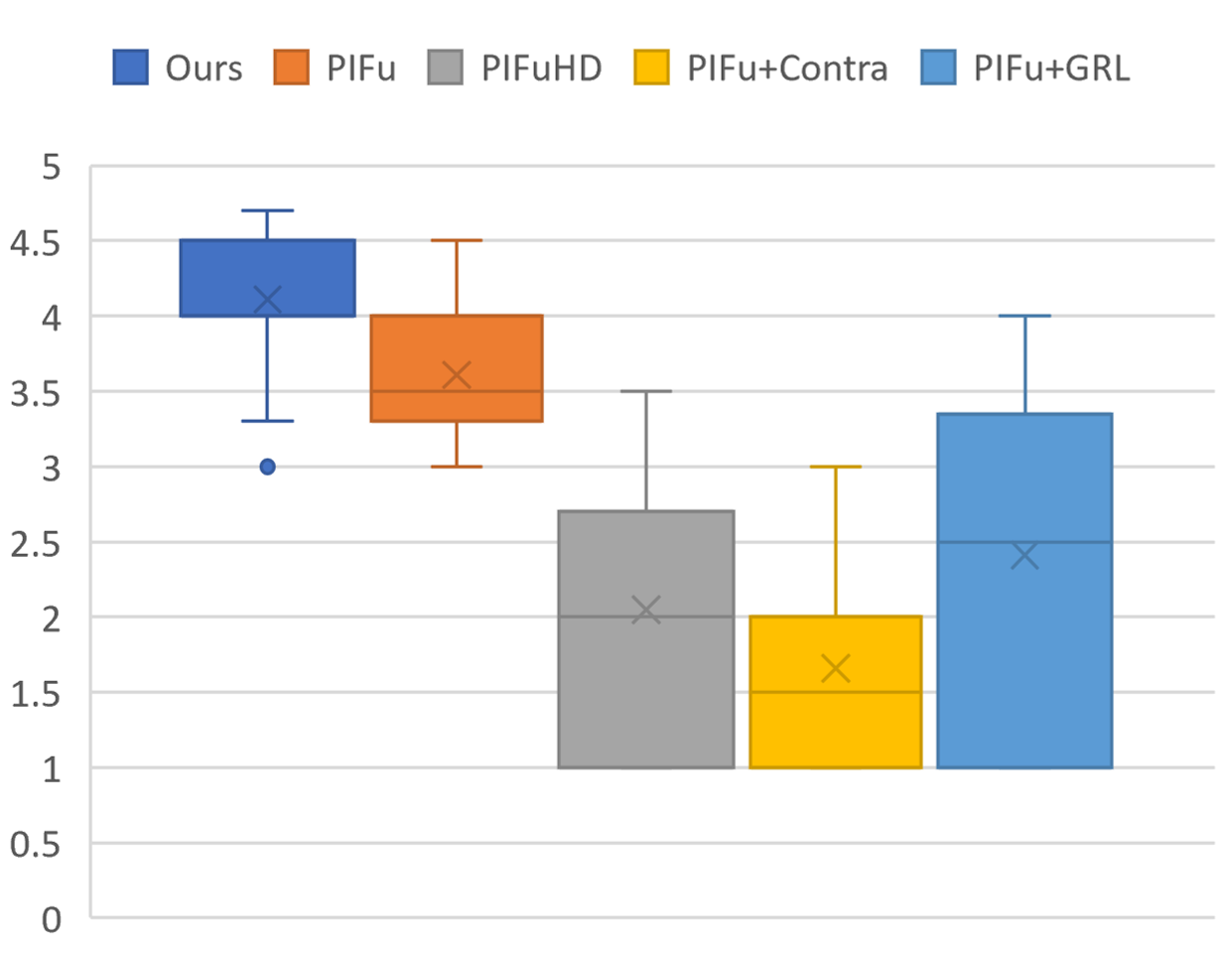}
  \caption{\label{fig:userstudy}%
           User study on results. We calculate the averaged score for results generated by each method.
           }
\end{figure}

To evaluate the results from the users' perspective, we design a user study. We invited 20 participants (10 \textcolor{black}{males} and 10 \textcolor{black}{females}, the mean age is 34.75 and the standard deviation is 16.98) coming from different educational/occupational backgrounds. Each participant was presented with one 3D shape and a corresponding RGB image at a time. To avoid bias, the results from various methods are mixed, unlabelled and shown in random order. We asked each participant to give a score from 1 to 5 (the higher, the better) based on the similarity of the 3D shape and RGB image. The results are presented in Figure \ref{fig:userstudy}. Our method achieves the highest score.

\subsection{VR Applications}

\begin{figure}[tbph]
  \centering
  \includegraphics[width=0.8\linewidth]{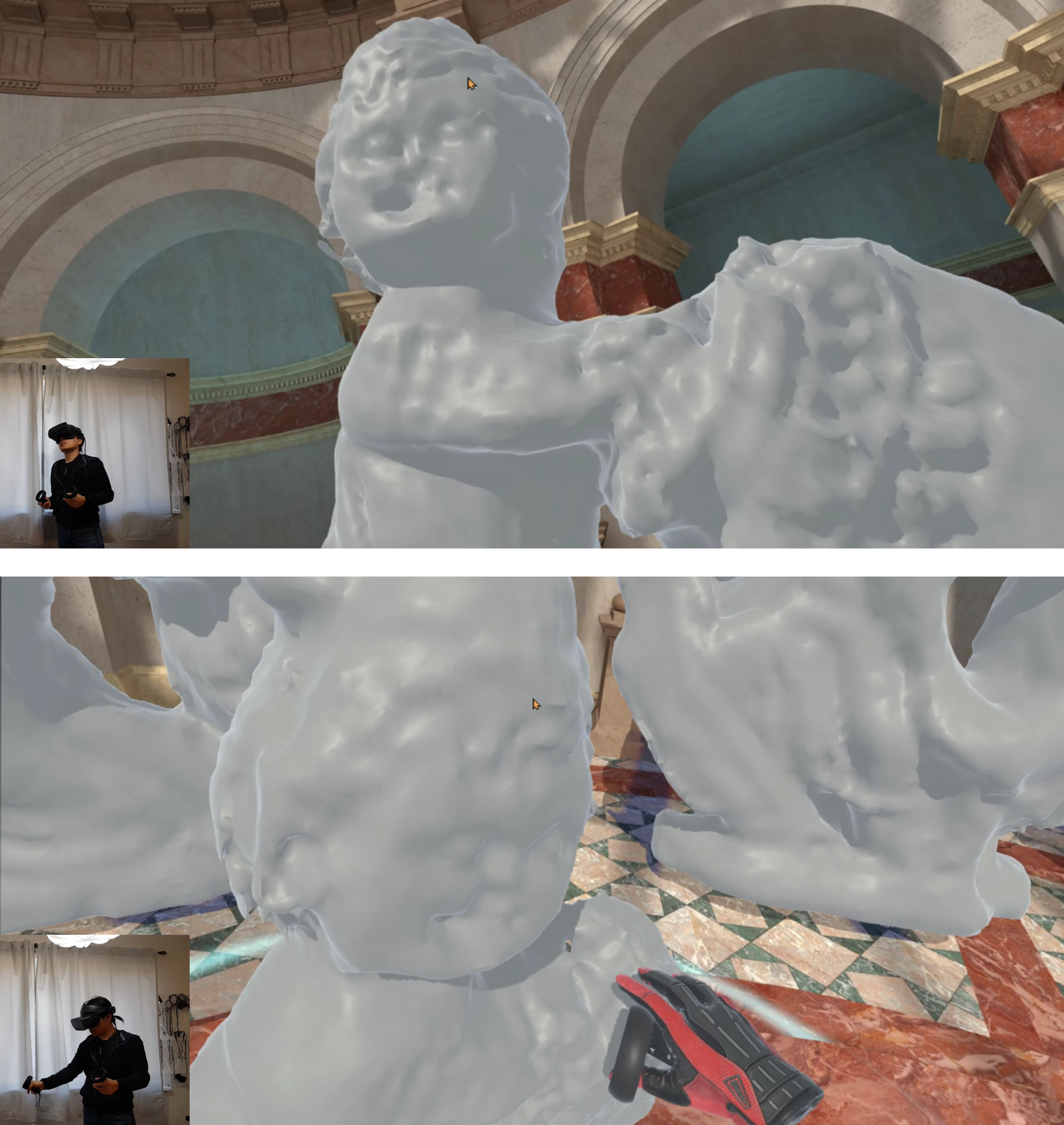}
  \caption{\label{fig:vr}%
           VR displaying of our results. (Top image) Seeing the generated sculpture. (Bottom image)  Locating the result.
           }
\end{figure}

We also display our results on a commercial VR headset. With VR equipment, users can place, move and observe the generated sculptures from all directions as shown in Figure \ref{fig:vr}.

\section{Conclusion}

In conclusion, we are the first to apply unsupervised domain adaptation into implicit models to solve the problem of reconstructing 3D sculptures from single-view RGB images, and our method can be deployed in creating various content for VR/AR applications. Our method addresses two critical challenges. The first one is the limited data acquisition, leading to inadequate data for supervised learning and incomplete shape information. The second one is the domain shift of irregular topology. We propose our unsupervised domain adaptation based on multi-level features with distribution alignment and neighbourhood information to transfer pre-trained PIFu to solve the problem of single-view sculpture reconstruction. Our adaptation method makes use of features from multiple layers of PIFu, combining and processing meaningful information from different levels. With multi-layer features, our adaptation explores the underlying structure of latent feature space and leverages pseudo labels to gradually obtain correct predictions on the target domain. Experiments have shown that our domain adaptation method can be successfully applied into PIFu. Compared with other methods without 3D ground truth, our method provides better performance measured by P2S and CD. Compared with supervised methods, our method generalizes better and our training does not require 3D ground truth. We find out that supervised methods are suffering with limited datasets. They cannot handle diverse surfaces. Our ablation studies validate the necessity of our proposed modules. Finally, we invited participants for a user study.

\begin{figure}[tbph]
  \centering
  \includegraphics[width=1.0\linewidth]{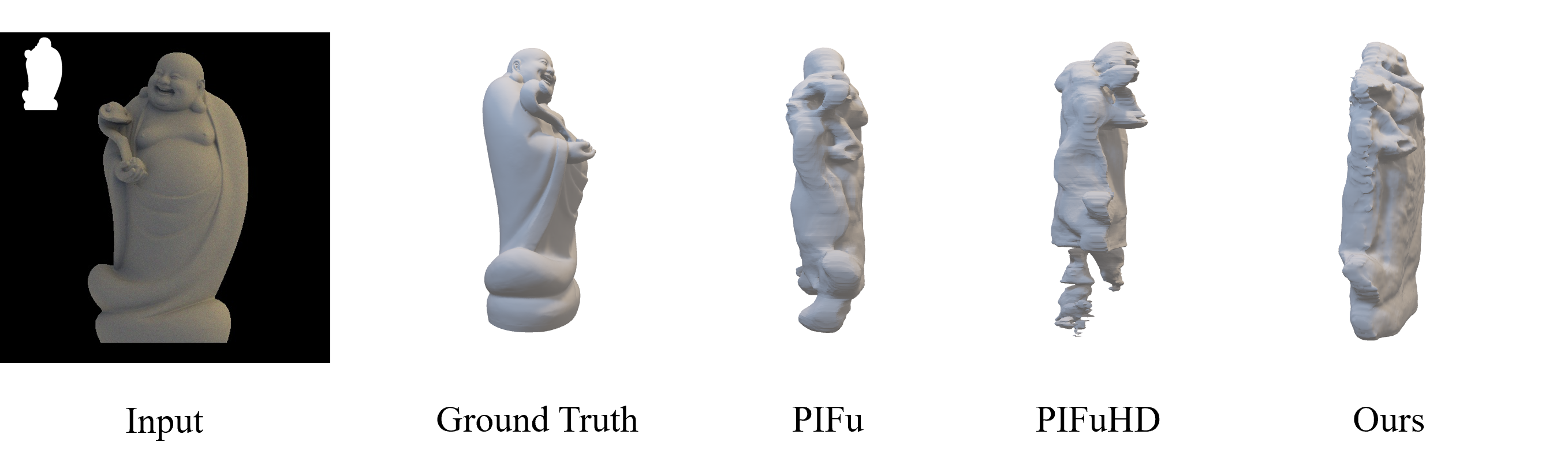}
  \caption{\label{fig:failure_case}%
           One challenging case in experiment. The network is not able to obtain the correct shapes for overweight people.
           }
\end{figure}

In the future, we will focus on three potential directions. One potential research direction is to apply our method on artistic human paintings. Considering our adaptation network does not require labels, it is highly feasible to operate on artworks where only a 2D image is accessible and 3D ground truth is not available at all. We intend to be able to handle more highly abstract human shapes such as Picasso paintings. It potentially is difficult for our adaptation method to find an appropriate mapping relationship. The other direction is treating challenging cases as shown in Figure \ref{fig:failure_case}. This may be because knowledge transfer is largely influenced by the pre-trained models. We will explore deeper structures of sculpture features to minimize such influence in the future. \textcolor{black}{Our domain adaptation has potential to be applied into other pretrained implicit-function models with point-features and no extra inputs. Generally, our method can be applied to a pretrained explicit reconstruction model. We will leave such attempts for future work.}

\bibliographystyle{ACM-Reference-Format}
\bibliography{main}

\end{document}